\def\year{2021}\relax
\documentclass[letterpaper]{article} % DO NOT CHANGE THIS
\usepackage{aaai21}  % DO NOT CHANGE THIS
\usepackage{times}  % DO NOT CHANGE THIS
\usepackage{helvet} % DO NOT CHANGE THIS
\usepackage{courier}  % DO NOT CHANGE THIS
\usepackage[hyphens]{url}  % DO NOT CHANGE THIS
\usepackage{graphicx} % DO NOT CHANGE THIS
\usepackage{float}
\usepackage{stfloats}
\usepackage{natbib}  % DO NOT CHANGE THIS AND DO NOT ADD ANY OPTIONS TO IT
\usepackage{caption} % DO NOT CHANGE THIS AND DO NOT ADD ANY OPTIONS TO IT
\usepackage[caption=false]{subfig}
\usepackage{mlmath}

\usepackage[switch]{lineno}

\title{Force-in-domain GAN inversion}
\author {
        Guangjie Leng \textsuperscript{\rm 1,}\thanks{Equal contribution.},
        Yekun Zhu \textsuperscript{\rm 1,*},
        Zhi-Qin John Xu \textsuperscript{\rm 1,2,}\thanks{Corresponding author.}\\
}
\affiliations {
    1 Institute of  Natural Sciences and School of Mathematical Sciences and  MOE-LSC, Shanghai Jiao Tong University\\
    2 Qing Yuan Research Institute, Shanghai Jiao Tong University  \\
    lengguangjie@sjtu.edu.cn, 
    zhuyekun123@sjtu.edu.cn, xuzhiqin@sjtu.edu.cn
}
\begin{document}
% \linenumbers
\maketitle

\begin{abstract}
Empirical works suggest that various semantics emerge in the latent space of Generative Adversarial Networks (GANs) when being trained to generate images. To perform real image editing, it requires an accurate mapping from the real image to the latent space to leveraging these learned semantics, which is important yet difficult. An \textit{in-domain} GAN inversion approach is recently proposed to constraint the inverted code within the latent space by forcing the reconstructed image obtained from the inverted code within the real image space. Empirically, we find that the inverted code by the \textit{in-domain} GAN can deviate from the latent space significantly. To solve this problem, we propose a \textit{force-in-domain} GAN based on the \textit{in-domain} GAN, which utilizes a discriminator to force the inverted code within the latent space. The \textit{force-in-domain} GAN can also be interpreted by a cycle-GAN with slight modification. Extensive experiments show that our \textit{force-in-domain} GAN not only reconstructs the target image at the pixel level, but also align the inverted code with the latent space well for semantic editing.
\end{abstract}

\section{Introduction\label{sec:Introduction}}
Generative Adversarial Networks (GANs) \citep{goodfellow2014generative} are successful in synthesizing high fidelity real images, such as face. A GAN consists of two key competing components, i.e., a generator part and a discriminator part. The generator network learns to map from a latent code, which is often sampled from a random distribution, to synthesized image. The discriminator network is trained to discriminate the real and the fake images, while the generator network is trained to make the synthesized image better to fool the discriminator. 

A series of works \citep{goetschalckx2019ganalyze, jahanian2019steerability,shen2020interfacegan} suggest that during the training, semantics spontaneously emerges within the latent code. The manipulation of the corresponding attributes of the latent code leads to change of the image in semantic. To enable the edit for real images, it is inevitable to obtain the inverted latent code for the real image. A simple minimization of the Mean Squared Error (MSE) between the generated samples and the associated real samples often leads to unsatisfied latent codes, because this approach ignores the domain constraints for both inverted codes and reconstructed images. Small MSE does not imply the reconstructed images lying in the real image space. \cite{zhu2020domain} propose an \textit{in-domain} GAN to force the reconstructed image lying in the real image space. The semantics editing based on 
the inverted code obtained by the \textit{in-domain} GAN is much better than the naive approach. However, our experiments show that the inverted codes obtained by the \textit{in-domain} GAN often deviate from the latent space significantly.  

In this work, we further rescue in-domain crisis for the latent code. Based on the \textit{in-domain} GAN, we propose a \textit{force-in-domain} GAN. We add a discriminator network for the inverted code, therefore, the inversion network can be trained to be good enough to fool the discriminator. The  \textit{force-in-domain} GAN works between two domains, i.e., the real image domain and the latent code domain, thus, it can also be interpreted by a cycle-GAN but with slight modification. Our experiments show that the inverted code obtained by the \textit{force-in-domain} GAN faithfully overlaps with latent space and reconstructs the target images at the pixel level and semantic level. In addition, extensive experiments show that our \textit{force-in-domain} GAN shows good performance for semantic editing.

% GANs are formulated as a two-player game between generator for synthesize images and a discriminator to differentiate real data from fake data. Based on such generative training, GANS learn to map a random distribution to the real data observation and the product photo-realistic images from randomly sampled latent codes.

% Recent work has shown that GANs spontaneously learn to encode rich semantics inside the latent space and that vary the latent codes leads to the manipulation of the corresponding attributes occurring in the output images. But how can we get the latent codes with capability of editing images remains a difficult problem.

% The existing approaches for addressing this issue fall into two categories. The common approach is called the GAN inversion that is based on the optimization of Mean Squared Error (MSE) between the generated samples and the associated real samples. This category don't take the semantics into consideration. The second category is named as adversarial inference, which infer the latent code by using another GANs, such as \textit{In-domain} GAN. Although the category  consider the semantics, the inverted codes of such category have a different distribution with real data, which makes real images editing meaningless.

%--------------- related works --------------------------
 \section{Related Works}\label{section2}
{\bfseries Generative Adversarial Networks.} GANs can learn the distribution of real images through adversarial training to generate realistic images \citep{goodfellow2014generative}.
% The emergence and use of GAN has greatly increased people's achievements in image generation speed and image generation quality. At the same time, recently various aspects of research have explored why GAN can generate high-quality pictures, what happens to GAN when generating pictures, and how to control it. 
It has been found that GANs spontaneously learn semantics in the latent space, which makes it feasible for controlling and explaining the generation process of GAN. In theory, \cite{chen2018metrics, arvanitidis2017latent, kuhnel2018latent} use Riemannian manifold to study the semantic editing of GANs, \cite{shen2020interfacegan} focus on GANs that generates face images, and explores the connection between semantic space and actual images. However, due to the limitation of the GANs' structure, it is still difficult to freely edit the semantics of the latent space to change the generated images.
\\
{\bfseries GAN Inversion.} GANs learn a map from a random distribution to the real data distribution. Hence, it is for making inference on real images. GAN inversion can help us apply the semantic editing of the latent space to the generated pictures. Given a fixed GAN model, GAN inversion aims at finding the most accurate latent code to recover the input images. The concept of GAN inversion is described in detail in \cite{perarnau2016invertible, lipton2017precise, creswell2018inverting}. Prior
work to achieve GAN inversion are mainly divided into two ways, one is to set an optimization problem (\cite{abdal2019image2stylegan, ma2019invertibility, abdal2020image2stylegan++}), and the other is to use the trained GAN generator to construct the encoder (\cite{dumoulin2016adversarially, donahue2016adversarial, zhu2019lia}).
\\
{\bfseries Semantic Faces Editing with GANs.} Semantic face editing aims to realize the manipulation of face attributes by manipulating the latent variables of the latent space. In semantic face editing, we hope that during the editing process, only the face attributes we operate have changed, while other face attributes remain unchanged. This is the difficulty and key point of semantic face editing. In order to achieve this goal, different methods are proposed: \cite{odena2017conditional, chen2016infogan, tran2017disentangled} constructs a special loss function to realize the semantic editing of human faces; \cite{donahue2017semantically, shen2018faceid} construct a special structure of GAN to achieve semantic editing.
\\

\section{\textit{Force-in-domain} GAN inversion}
 \begin{figure*}[hb]
     \centering
     \includegraphics[width=0.8\textwidth]{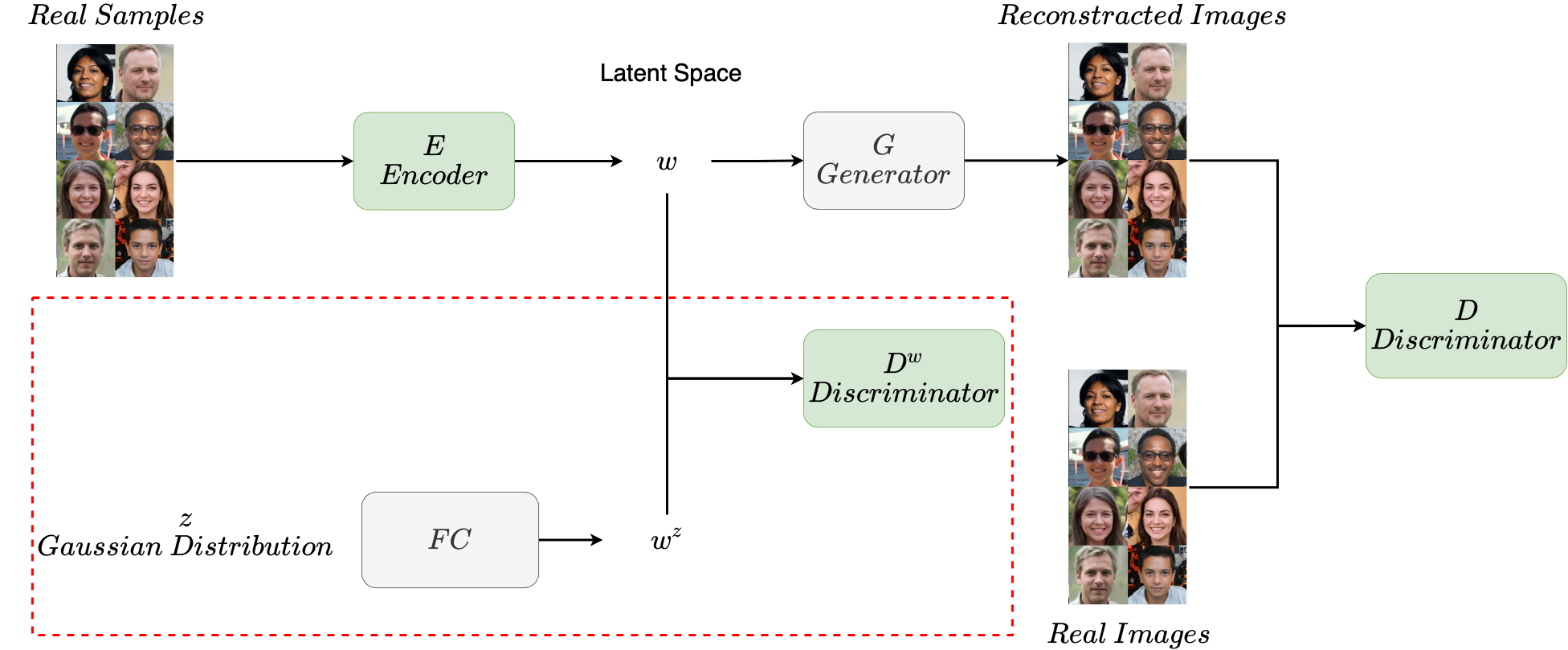}
     \caption{Our \textit{force-in-domain} GAN inversion structure. Note that FC block is from the SyleGAN that maps from a Gaussian distribution to the latent space. Only green blocks are trainable.}
     \label{fig:structure}
 \end{figure*}
In GAN model, the semantics
 refer to the emergent knowledge that GAN has learned from the observed data. Style-GAN model generates realistic images based on a disentangled representation in the  latent space $\mathcal{W}$, which is obtained by samples from a Gaussian distribution through a fully-connected neural network \citep{karras2019style}. The latent space $\mathcal{W}$ is interpreted as a disentangled space of different image styles. A good GAN inversion model should not only  recover input image by the pixel values, but also account for inverting the latent code semantically. For this purpose, we propose a \textit{force-domain-guided} encoder which forces the inverted latent code having the same distribution with the latent space in StyleGAN. In general, the distribution of the latent space is unknown. To force the inverted latent code align with the original latent space, a new discriminator is added in our model, compared with the in-domain GAN  \citep{zhu2020domain}, as shown in  Fig. \ref{fig:structure}.

 \subsection{\textit{Force-in-domain} GAN structure} 
%   {\bfseries Choice of the latent space} Typically, GANs maps sample $\mathbf{z}$ from the Gaussian distribution to images. However, StyleGAN model \citep{karras2019style} first uses a fullly-connected network FC to map $z$ to the latent code $w^{z}$, which is interpreted as a disentangled space of different image styles, then, a generator maps the latent code $w^{z}$ to images $x$. In our work, similarly to \cite{zhu2020domain}, we train encode to maps from images to latent code $w$.
  
%  {\bfseries Problem Statement.} 
 The \textit{in-domain} GAN consists of the following components: encoder, $E(\cdot):\mathcal{X}\rightarrow\mathcal{W}$, obtains the latent code corresponding to the input image; generator, $G(\cdot):\mathcal{W}\rightarrow\mathcal{X}$, synthesizes high-quality images; discriminator $D(\cdot)$ distinguishes real data from synthesized data. Our \textit{force-in-domain} GAN adds the following components: the fully connected layer FC from the StyleGAN that maps from the Gaussian distribution $z$ to the latent space $w^{z}$; a new discriminator $D^{w}(\cdot)$ distinguishes inverted latent codes from real distribution of the latent space. Encoder $E$ is the reverse mapping of the generator $G(\cdot)$, i.e., mapping from images $\mathbf{x}^{real}$ to latent code $\mathbf{w}$ that can recover the real images $\mathbf{x}^{real}$. Our purpose is to  align $w$ with $w^{z}$ of the prior knowledge in the pre-trained StyleGAN model. The structure of the \textit{force-in-domain} GAN is depicted in Fig. \ref{fig:structure}. Note that only green blocks are trainable.

%  for GAN inversion task\citep{abdal2019image2stylegan,karras2020analyzing,zhu2019lia}. \textit{In-domain-GAN} proposed three reasons for choosing $\mathcal{W}$ space as the inversion space \citep{zhu2020domain}. In our work, we conduct all experiments in the $\mathcal{W}$ space, but our approach can be performed in the $\mathcal{Z}$ space as well.
 
 \subsection{Interpretation by Cycle-GAN}
% Cycle-gan has achieved great success in style translation， which can guarantee the translation “cycle consistent”, in the sense that if we translate, e.g., a sentence from English to French, and then translate it back from French to English, we should arrive at the original sentence. Mathematically, if we have a translator G:X$\rightarrow$Y and another translator F: Y$\rightarrow$X, then G and F should be the inversion of each other, and both mapping should be bijections. This structure can be achieved by training G and F simultaneously, and adding a cycle consistency loss that. Encourage F(G(x)) $\sim$ x and G(F(y)) $\sim$ y. Combing this loss with adversarial losses on domains X and Y yields our full objective for unpaired image-to-image translation.
\begin{figure*}
    \centering
    \includegraphics[width=0.8\textwidth]{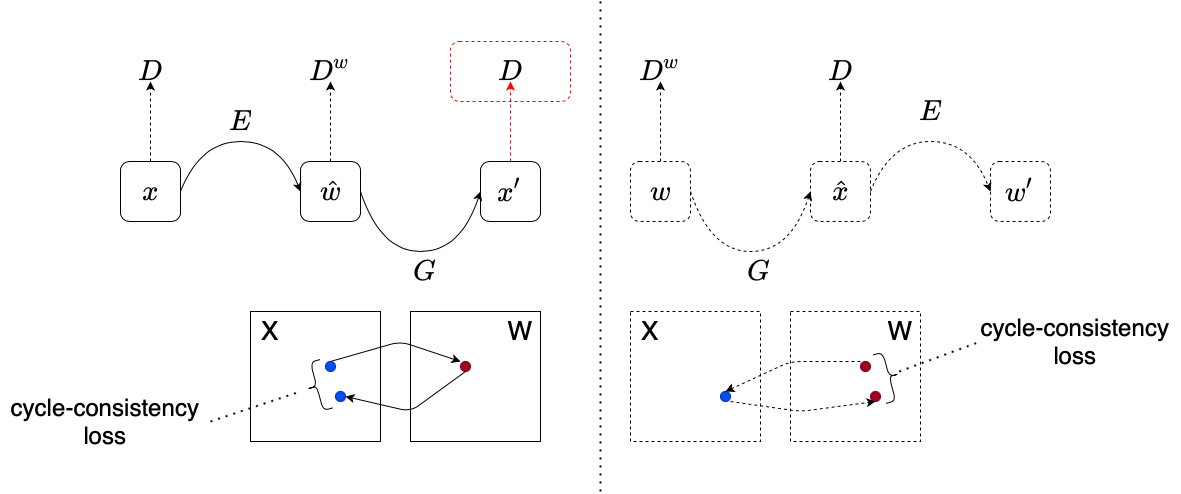}
    \caption{Interpretation by Cycle-GAN: the whole structure without the red block is the cycle-GAN. Our \textit{force-in-domain} GAN is the left part with the red block and the block is the new discriminator we introduce in our network.}
    \label{cycle-gan}
\end{figure*}

As shown in Fig. \ref{cycle-gan}, cycle-GAN (without the red dashed block)  transfers variables $x$ in one domain to $w$ in the other domain by $E$ and also converts $w$ back to $x$ by $G$. In the context of \textit{force-in-domain} GAN, two domains are the image domain and latent space, respectively. In Cycle-GAN, two discriminator networks are imposed to force $\hat{w}$ and $\hat{x}$ in their  domains, respectively.

In our work, the generator $G$ is already well-trained, and we only focus on the reconstruction of real images, therefore, \textit{force-in-domain} GAN can be regarded as left part of cycle-GAN. In addition, a domain discriminator $D$ (in the red dashed block) is equipped for the reconstructed image $x^{\prime}$ to ensure $x^{\prime}$ in the real image domain.

 \subsection{Adversarial Loss}
 We apply adversarial losses  \citep{goodfellow2014generative} to the two discriminators. The   objective is:
 \begin{equation}
\begin{aligned}
\mathcal{L}_{adv}^{D} &=
\mathop{\mathbb{E}}\limits_{\mathbf{x}^{real}\sim P_{data}}[\log(D(\mathbf{x}^{real}))]\\
&+\mathop{\mathbb{E}}\limits_{\mathbf{x}^{real}\sim P_{data}}[\log(1-D(G(E(\mathbf{x}^{real}))))] \\
& +\frac{\gamma}{2}\mathop{\mathbb{E}}_{\mathbf{x}^{real}\sim P_{data}}[\Vert\nabla_{\mathbf{w}}D(\mathbf{x}^{real})\Vert^2],
\end{aligned}
\end{equation}
\begin{equation}
\begin{aligned}
\mathcal{L}_{adv}^{D^{w}} &=
\mathop{\mathbb{E}}\limits_{z\sim \mathcal{N}(0, \mathbf{I})}[\log(D^{w}(FC(z)))]\\
&+\mathop{\mathbb{E}}\limits_{x^{real}\sim P_{data}}[log(1-D_{\mathcal{W}}(E(x^{real})))] \\
& +\frac{\gamma}{2}\mathop{\mathbb{E}}_{w\sim FC(\mathcal{N}(0, \mathbf{I}))}[\Vert\nabla_{\mathbf{w}}D^{w}(w)\Vert^2],
\end{aligned}
\end{equation}
where $\gamma$ is the hyper-parameter for the gradient regularization, $P_{data}$ is the distribution of the real image data and $\mathcal{N}(0, \mathbf{I})$ denotes the Gaussian distribution.
\subsection{Forward Cycle Consistent Loss}
In GAN inversion tasks, it requires the reconstructed images close to the original images, i.e. $\mathbf{x}\rightarrow E(\mathbf{x})\rightarrow G(E(\mathbf{x}))\approx \mathbf{x}$. This forward cycle consistent loss can be expressed as follows:
\begin{equation}
\mathcal{L}_{cyc} = \mathop{\mathbb{E}}\limits_{\mathbf{x}^{real}\sim P_{data}}\Vert\mathbf{x}^{real} - G(E(\mathbf{x}^{real}))\Vert_{2}.
\end{equation}
\subsection{Perceptual Loss}
Previous work has shown that high-quality images can be generated by defining and optimizing perceptual loss functions based on high-level features extracted from pre-trained networks  \citep{johnson2016perceptual}. In our paper, we choose VGG \citep{simonyan2014very} as the pre-trained networks to introduce the perceptual loss:
\begin{equation}
    \mathcal{L}_{vgg} = \mathop{\mathbb{E}}\limits_{\mathbf{x}^{real}\sim P_{data}}\Vert F(\mathbf{x}^{real}) - F(G(E(\mathbf{x}^{real})))\Vert_{2},
\end{equation}
where F($\cdot$) denotes the VGG feature extraction model.
\subsection{Force-in-domain Loss}
We train the \textit{force-in-domain} GAN, which is illustrated in Fig. \ref{fig:structure}, with the loss consisting of the forward cycle consistent loss, adversarial loss and the perceptual loss. By using these three kinds of losses, the encoder can inverse the input images to the   latent space $\mathcal{W}$ and reconstruct the original image in both the pixel level and the semantic level.  
\begin{equation}
\begin{aligned}
\mathcal{L}_{adv}^{E} = \mathcal{L}_{cyc} + \lambda_{adv}\mathcal{L}_{adv}^{D} + \lambda_{adv}^{D^{w}}\mathcal{L}_{adv}^{D^{w}} + \lambda_{vgg}\mathcal{L}_{vgg},
\end{aligned}
\end{equation}
where $\lambda$'s are hyper-parameters.
% where $P_{data}$ denotes the distribution of real data, $\mathcal{N}(0, I)$ denotes the Gaussian distribution and $\gamma$ is the hyper-parameter for the gradient regularization. $\lambda_{vgg}$, $\lambda_{data}$ and $\lambda_{adv}^{D_{w}}$ are the perceptual, the discriminator and the new discriminator loss weights. And F($\cdot$) denotes the VGG feature extraction model.
\section{Experiments}
In this section, we experimentally show that the latent codes obtained by the \textit{force-in-domain} GAN inversion overlaps well with latent space of the original StyleGAN. We also utilize the \textit{in-domain} GAN to edit real images. 
\subsection{Experimental settings}
We conduct experiments on FFHQ dataset\citep{karras2019style}, which contains 70,000 high-quality face images. The generator  $G$ to be inverted is from the pre-trained StyleGAN \cite{karras2019style}. When training the \textit{in-domain} GAN, the generator $G$ and the $FC$ components are fixed and we update the encoder $E$, the discriminator $D$ and $D^{w}$.
We set $\lambda_{vgg} = 5\times e^{-5}, \lambda_{adv} = \lambda_{adv}^{D^{w}} = 0.1$ and $\gamma = 10$.
\subsection{High quality of image reconstruction}
\begin{table}[ht]
    \centering
    \caption{Quantitative Comparison between different inversion methods. For each model, we invert 10k images for evaluation. $\downarrow$ means lower is better.}
    \begin{tabular}{|l|c|c|}
    \hline
    Method  &  FID$\downarrow$ & MSE$\downarrow$\\
    \hline
    In-Domain GAN Encoder without BN & 17.55 & 0.49\\
    \hline
    Force-in-domain Encoder &15.06 & 0.067\\
    \hline
    \end{tabular}
    \label{quantitative_comparison}
\end{table}
A necessary  condition for a good GAN inversion is that the reconstructed image is close to original image in both pixel and feature levels. To show that the \textit{force-in-domain} GAN can faithfully reconstruct the original image, we quantitatively compute two common indexes, that is, Frchet Inception Distance (FID) \citep{heusel2017gans} and Mean-Square Error (MSE). The FID is a metric for assessing the quality of images generated by GAN, which compares the distribution of generated images with the distribution of real images that were used to train the generator. MSE quantifies the difference between the reconstructed images and the original images in the pixel level. We randomly select $10,000$ images from FFHQ dataset for computation. As shown in Table
\ref{quantitative_comparison}, both FID and MSE of the \textit{force-in-domain} GAN are much smaller that those of \textit{in-domain} GAN. Therefore, the \textit{force-in-domain} GAN can improve the fidelity of the image reconstruction. Next, we would show in the latent space, the \textit{force-in-domain} GAN can also align the inverted latent codes well with the original latent space, thus, serving as a good model for real images editing due to the preservation of the semantics in latent space. 

\subsection{Image inversion}
\begin{figure}[hb]
    \centering
    \includegraphics[width=0.45\textwidth]{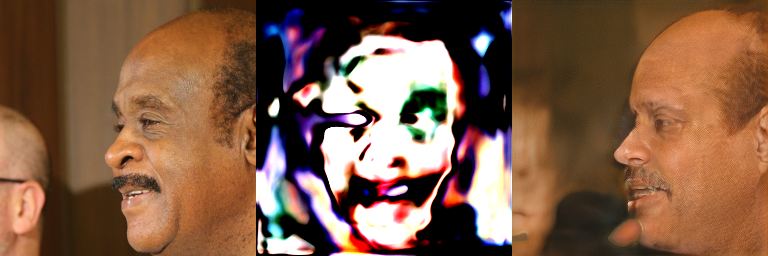}
    \includegraphics[width=0.45\textwidth]{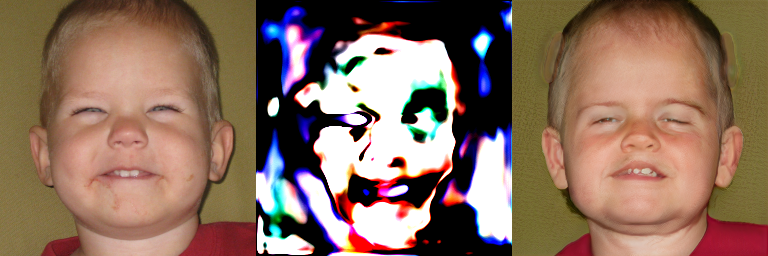}
    \caption{Reconstruction of images from FFHQ. Images in the left are the original. Images in the middle are the reconstruction results of \textit{in-domain} GAN. Images in the right are the reconstruction results of \textit{force-in-domian} FA}
    \label{fig:their_bad}
\end{figure}

We found that in the process of image reconstruction, \textit{in-domain} GAN would produce some strange images shown in Fig. \ref{fig:their_bad}. For these images, our \textit{force-in-domain} GAN can reconstruct well. Note that in our experiments, \textit{force-in-domain} GAN never produce such strange images. \citep{zhu2020domain} proposed BN-\textit{in-domain} GAN recently to solve this problem, but our further projection results also show that the reconstructed latent code in BN-\textit{in-domain} GAN is not on its real manifold, either. 
\subsection{Good alignment of latent codes}
\begin{figure}[hb]
    \centering
    \includegraphics[width=0.5\textwidth]{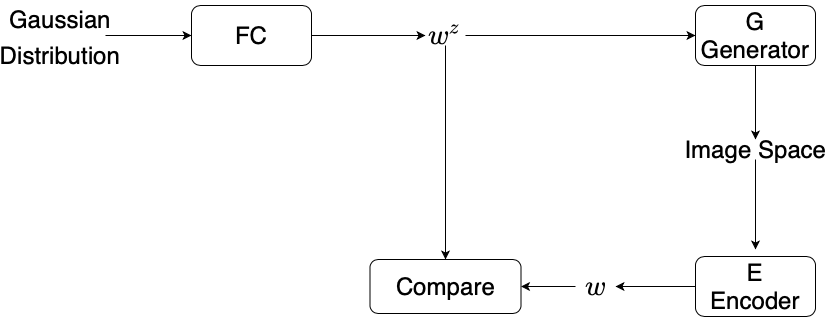}
    \caption{The idea of our projection method}
    \label{projection}
\end{figure}

It is reasonable that the high quality of image reconstruction is due to the good inverted latent codes. Compared with the \textit{in-domain} GAN, the \textit{force-in-domain} GAN makes the inverted latent code closer to the original latent space in the sense of distribution. To verify this point, we project high-dimensional latent codes into two-dimensional space. The procedure to perform this comparison is depicted in Fig. \ref{projection}. The original latent code $w^{z}$ is obtained through the sampling from Gaussian distribution and the FC component.  We pass $w^{z}$ through generator $G$ to obtain a image and then use the encoder $E$ to obtain the inverted latent code $w$. Then, we project a bunch of $w^{z}$'s and $w$'s into two-dimensional space by two methods, i.e., principle component analysis (PCA) and t-SNE \citep{van2008visualizing}. For both methods, we sample $100,000$ samples from the Gaussian distribution to ensure enough precision.

\begin{figure}[!hb]
    \centering
    \subfloat[Original projection results]{\includegraphics[width=0.25\textwidth]{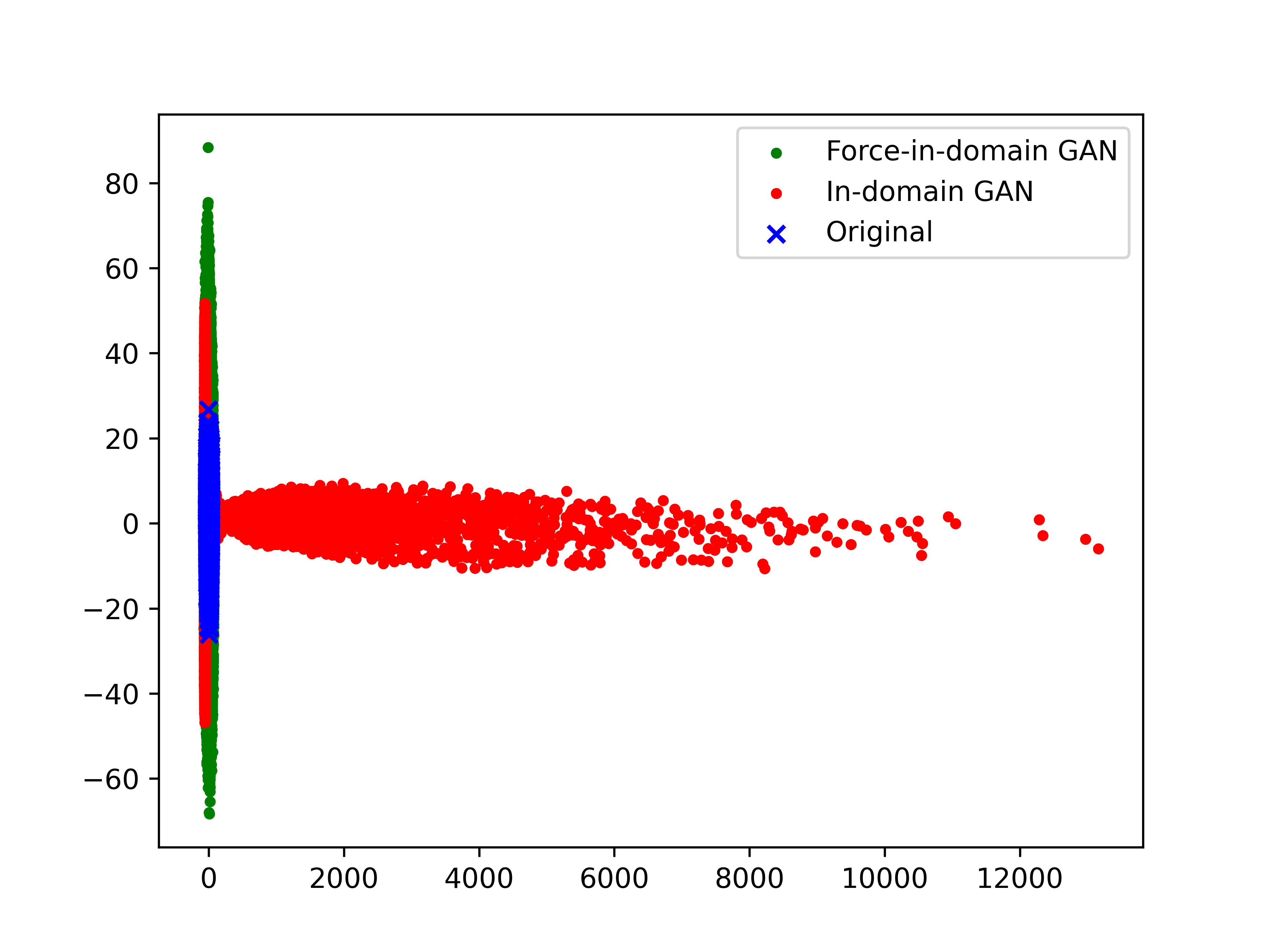}}
    \subfloat[Projection results without outliers]{\includegraphics[width=0.25\textwidth]{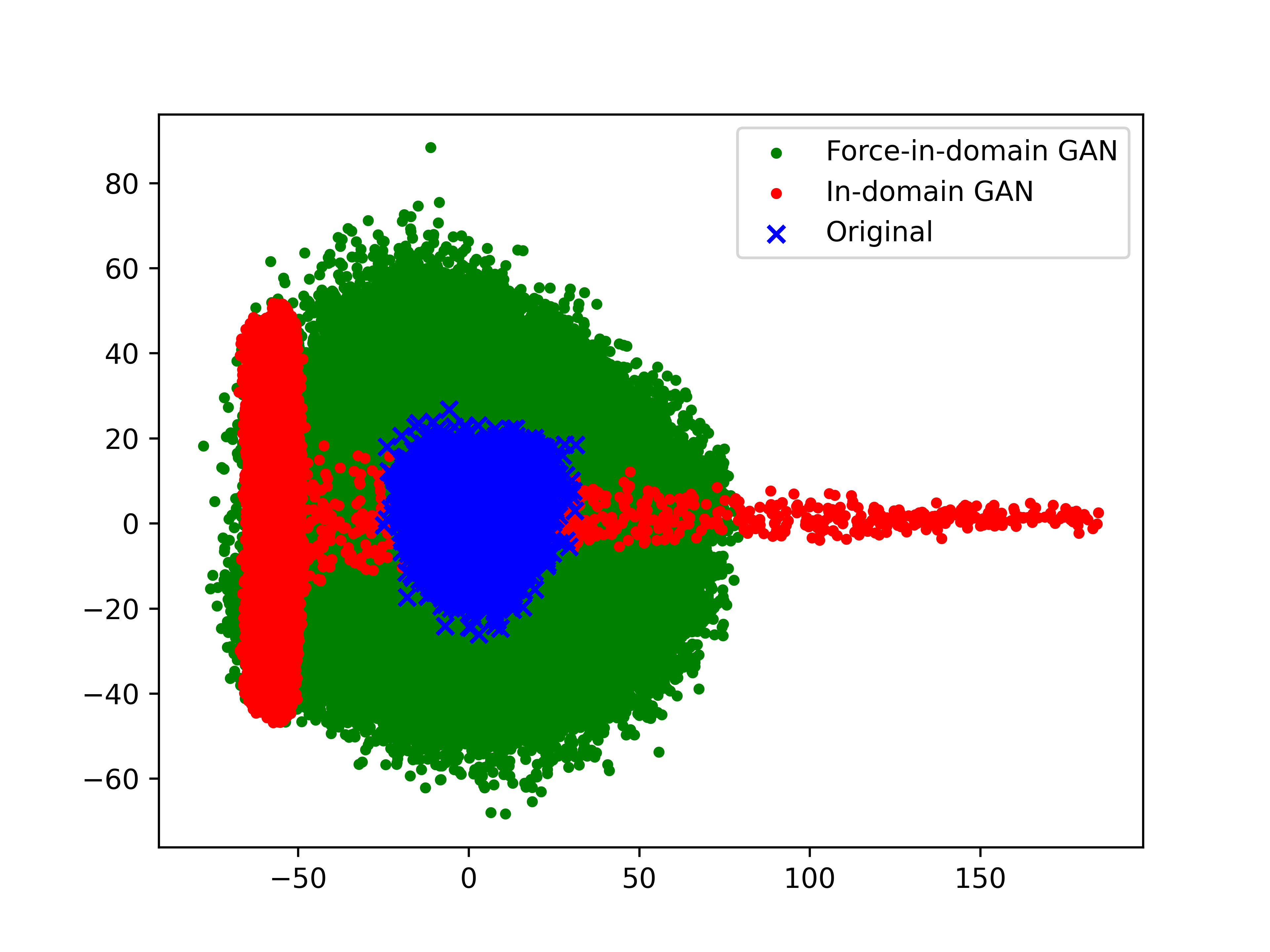}}
    \caption{Projection results using PCA method. In both figures, ``Original'' can the projection results of the true distribution of the   latent codes (sampling from Gaussian distribution and mapping to the   space); ``\textit{Force-in-domain} GAN'' means the projection results of \textit{force in-domain} GAN and ``In-domain GAN'' means the projection results of \textit{in-domain} GAN. Further more, both of the results of force-in-domian GAN and in-domain GAN are derived by the same samples.}
    \label{fig:pca_original}
\end{figure}

% The main idea of our projection shown in Fig. \ref{projection} is that we can obtain the intermediate latent codes by the encoders for real images, then project the obtained latent codes and the real codes together into the two-dimensional space. However, we don't know the real latent codes for arbitrarily given images. To solve this problem, we sample from Gaussian distribution first and then we use a pre-trained GAN model to synthesize face images. On the other hand, the mapping layer of the GAN can map those samples to the intermediate latent space, which can be viewed as the real distribution of the intermediate latent codes. After that, we can utilize the pre-trained encoders of \textit{in-domain} GAN and our model to attain the intermediate latent codes. Finally, the inverted latent codes and the true latent codes can be projected to $\mathbb{R}^{2}$ so that we can compare the distributions. In our paper, we have used two kinds of projection methods: PAC and t-SNE. For different methods, we sample 100000 samples from the Gaussian distribution so that the results of projection can be as precise as possible. 
The projection results of PCA method are shown in Fig. \ref{fig:pca_original}(a). There are so many outliers of the \textit{in-domain} GAN that we cannot visualize the comparison clearly. For visualization, we remove part of the data of \textit{in-domain} GAN, whose absolute values are bigger than ten times of the original projection results. As shown in Fig. \ref{fig:pca_original}(b), the projection of the inverted latent codes of \textit{in-domain} GAN deviate significantly from the true distribution, and there are many outliers that are far away from the true distribution. The projection results of \textit{force-in-domain} GAN are relatively concentrated and can cover the true distribution.
% And for their latest model, we have done similar experiments. 

\begin{figure}[h]
    \centering
    \subfloat[Original projection results]{\includegraphics[width=0.23\textwidth]{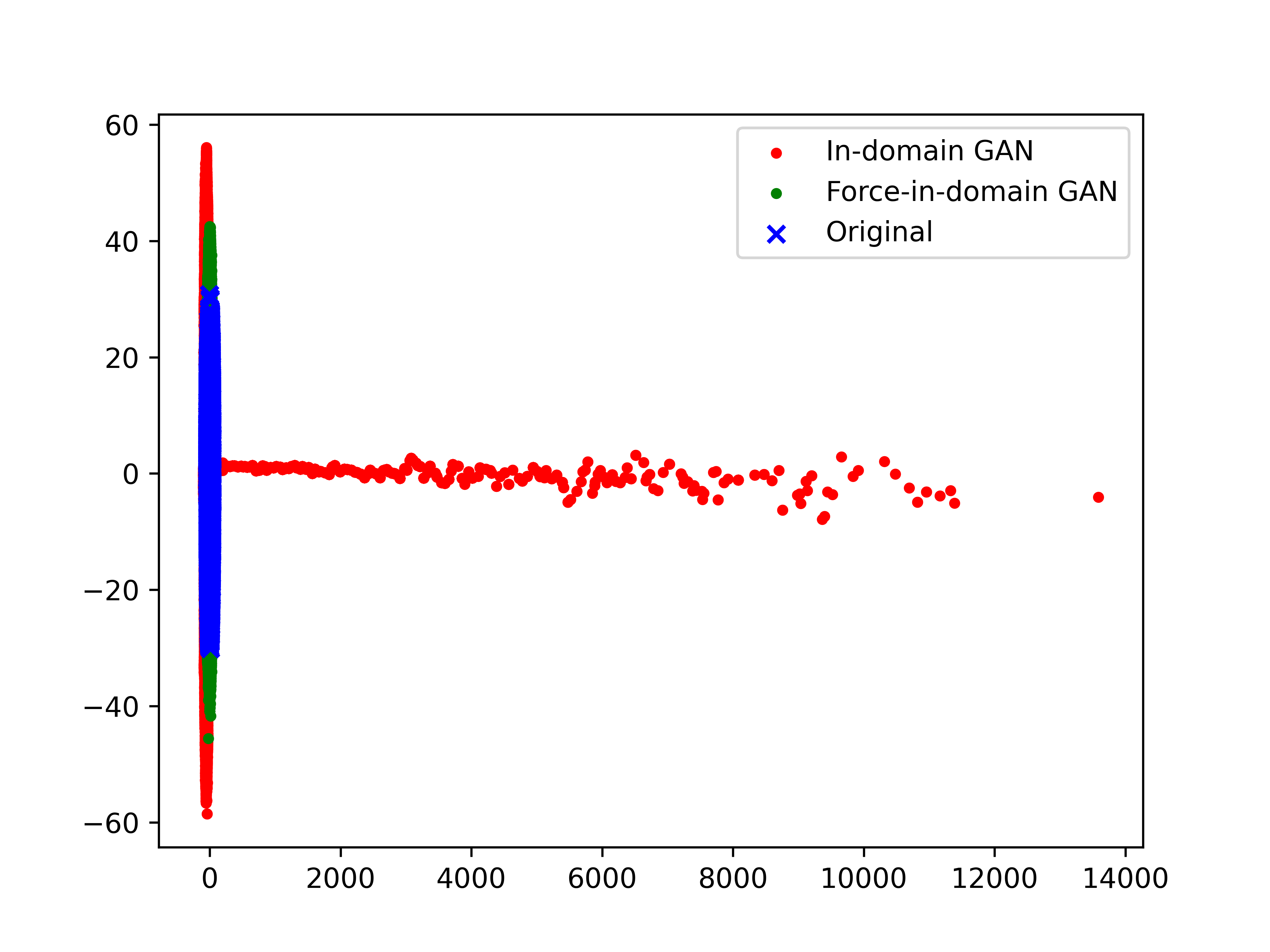}}
    \subfloat[Projection results without outliers]{\includegraphics[width=0.23\textwidth]{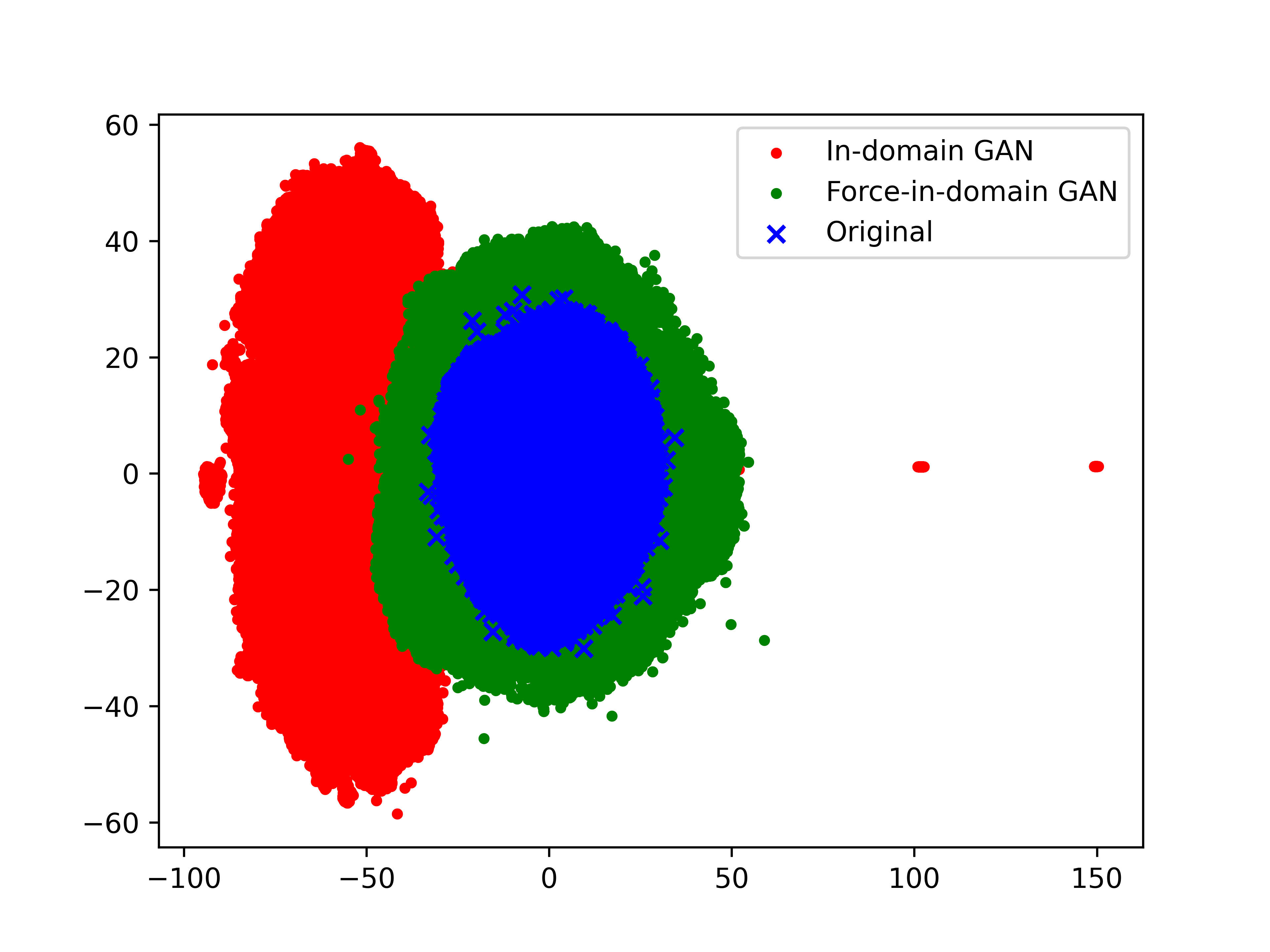}}
    \caption{Projection results using t-SNE method. Illustration is the same as Fig. \ref{fig:pca_original}.}
    \label{fig:tsne_original}
\end{figure}

\begin{figure}[hb]
    \centering
	\subfloat[Original projection results]{
	    \includegraphics[width=0.15\textwidth]{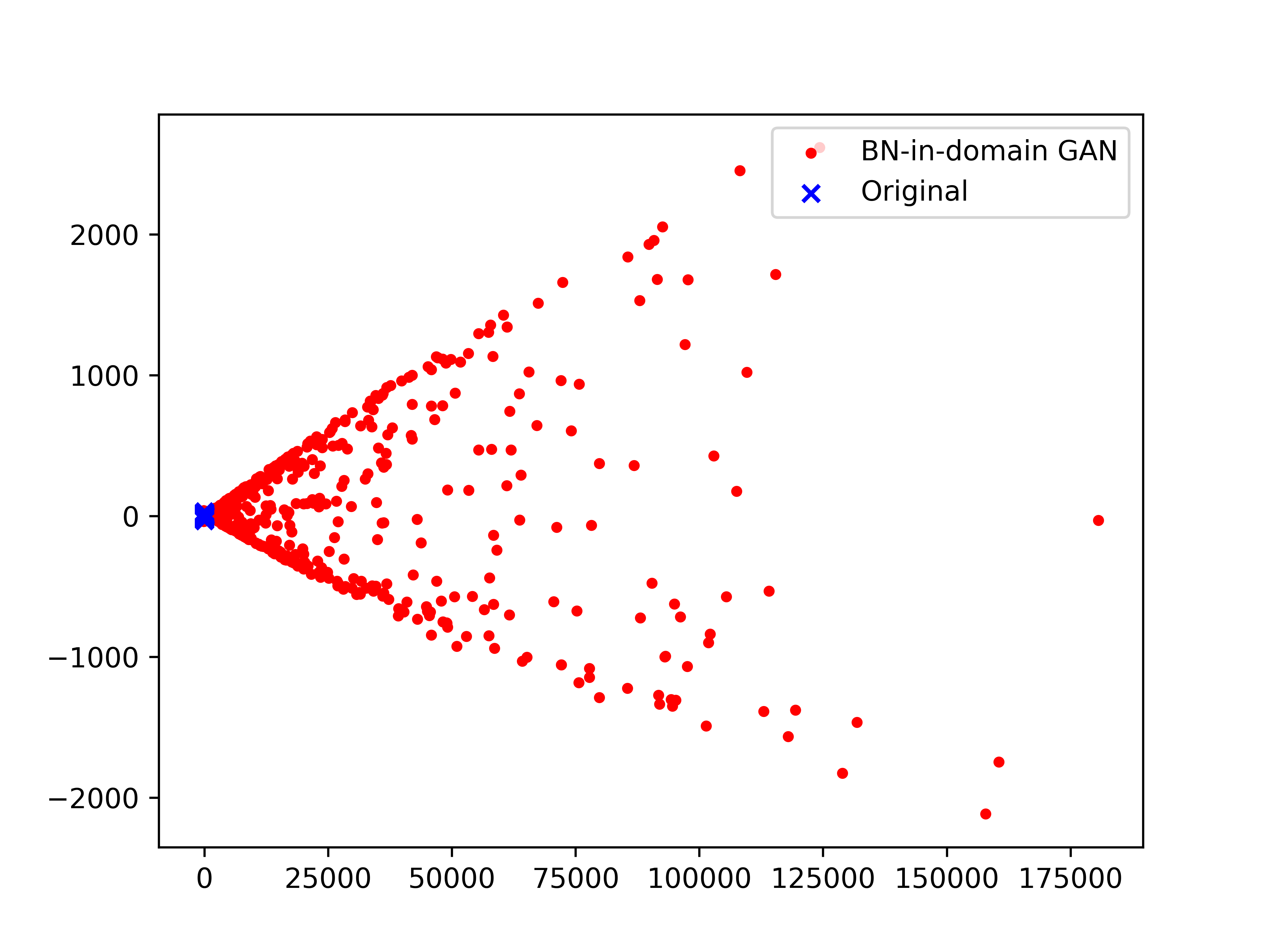}}
    \subfloat[Projection results without outliers]
        {\includegraphics[width=0.15\textwidth]{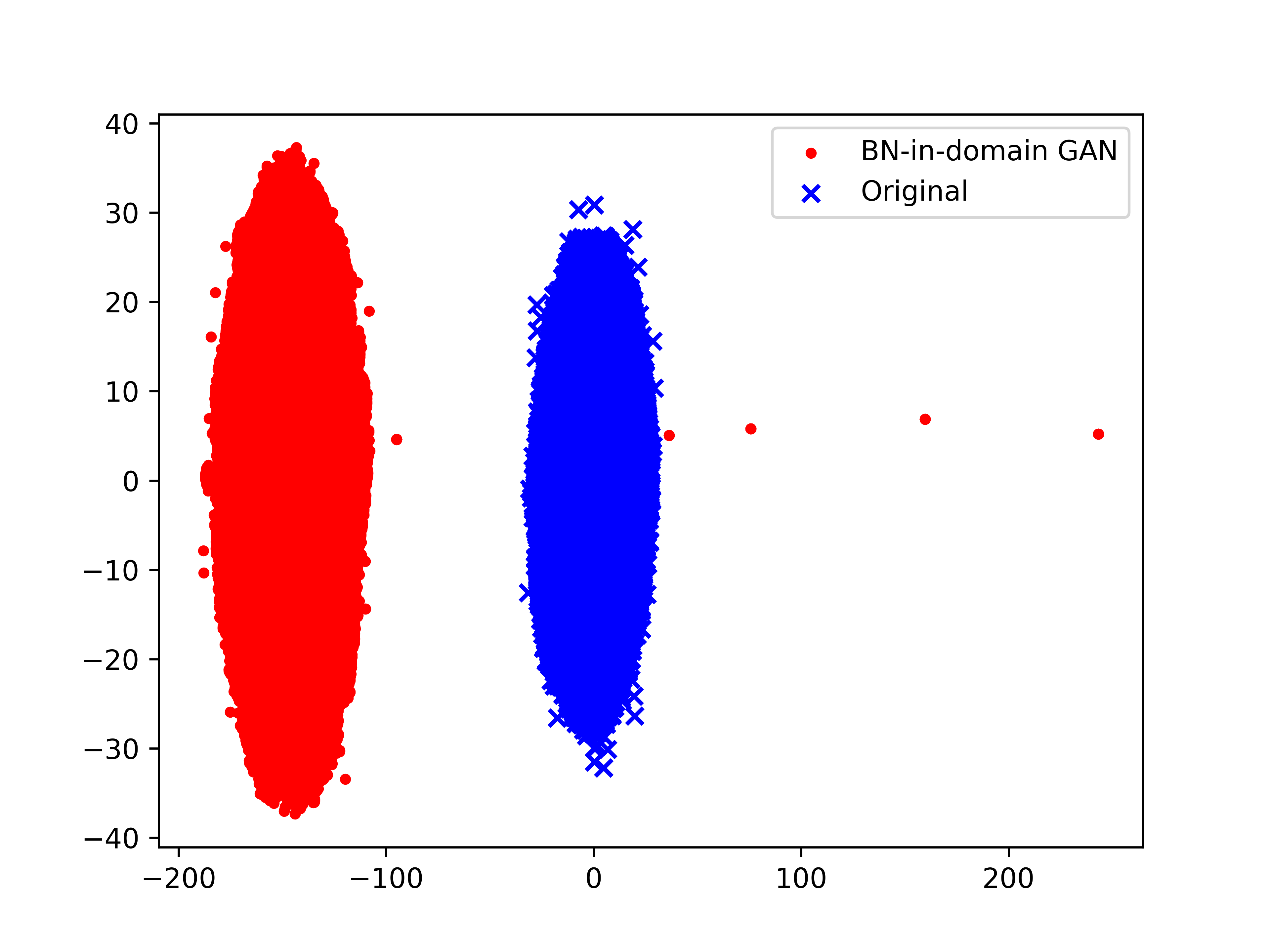}}
     \subfloat[Original projection results of our model]
        {\includegraphics[width=0.15\textwidth]{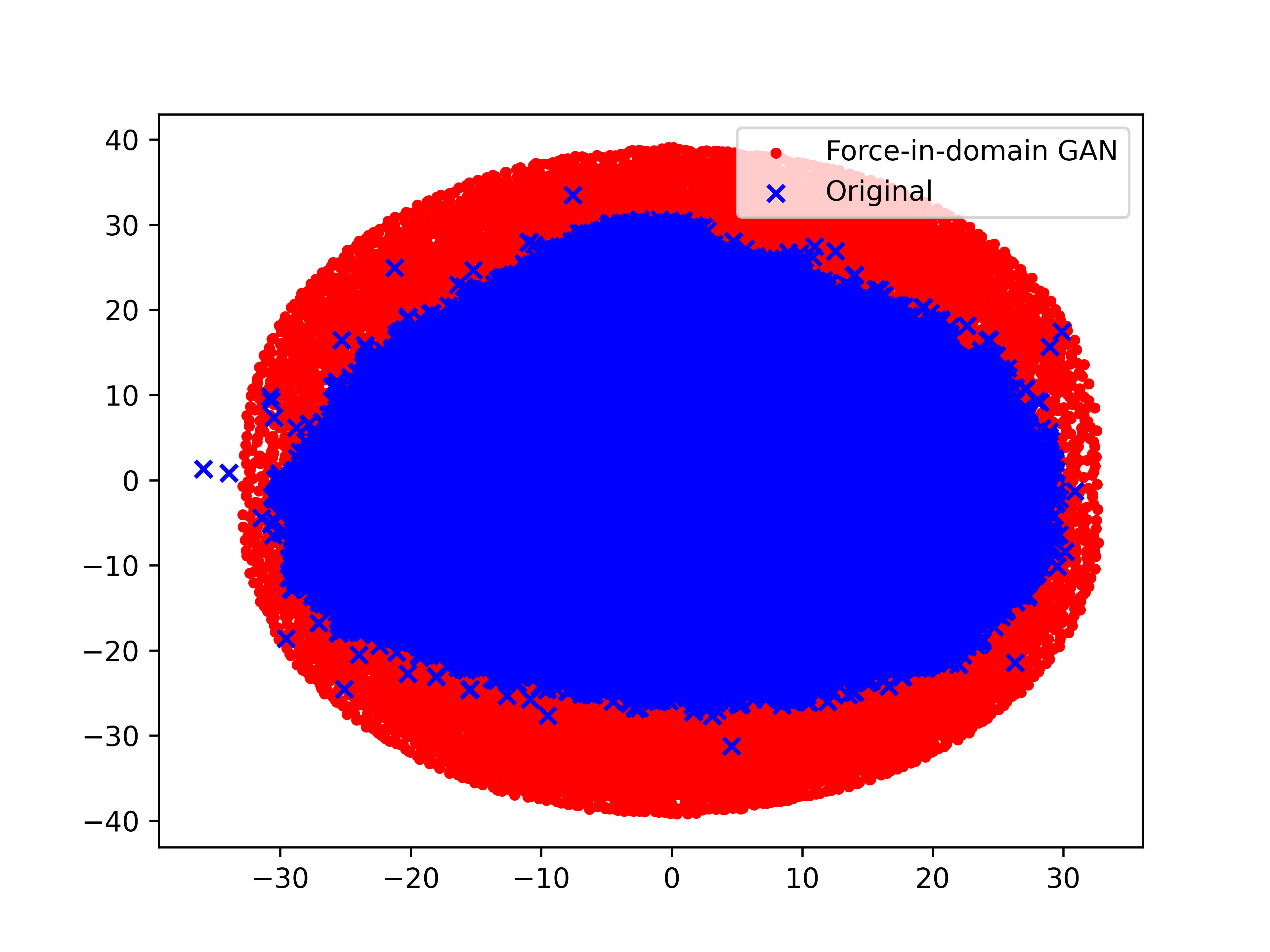}}
	\caption{Projection results using t-SNE method for \textit{BN-in-domain} GAN in (a,b) and  \textit{force-in-domain} GAN in (c). Illustration is the similar as Fig. \ref{fig:pca_original}.}
	\label{fig:tsne_update}
\end{figure}

For the projection results of t-SNE method, as shown in Fig. \ref{fig:tsne_original}(a), the projection of the \textit{in-domain} GAN also has many outliers. We similarly remove outliers and as shown in  Fig. \ref{fig:tsne_original}(b), the projection of the \textit{in-domain} GAN is significantly different from the projection of original latent codes. However, the projection of the \textit{force-in-domain} GAN overlaps with the projection of original latent codes much better. 
While preparing this paper, \cite{zhu2020domain} release a better  \textit{in-domain} GAN by adding batch normalization. For convenience, we denote this model by \textit{BN-in-domain} GAN.  We perform similar projection examination for this \textit{BN-in-domain} GAN. Since the FC components in \textit{BN-in-domain} GAN and \textit{force-in-domain} GAN are different, the original latent codes of these two models are different for a same Gaussian sample. We display the comparison of the inverted codes and the original latent codes for the \textit{BN-in-domain} GAN in Fig. \ref{fig:tsne_update}(a,b) and for the \textit{force-in-domain} GAN in Fig. \ref{fig:tsne_update}(c), respectively. Similarly, the inverted codes of the \textit{force-in-domain} GAN align well with the original latent codes while the \textit{BN-in-domain} GAN fails to align the inverted latent codes with the original latent space.  
From the projection results, we can find that our method can force the encoder to map real images much closer to the true latent space, thus, explaining the well reconstruction of image and being a better model for semantics editing for real images.
% \begin{figure}
% \centering
% \begin{minipage}[t]{0.3\textwidth}
% \centering
% \includegraphics[width=5cm]{figure/projection/tsne_update/tsne_in_domain_gan_100000_with_abnormal.png}
% \end{minipage}
% \begin{minipage}[t]{0.3\textwidth}
% \centering
% \includegraphics[width=5cm]{figure/projection/tsne_update/tsne_in_domain_gan_100000.png}
% \end{minipage}
% \begin{minipage}[t]{0.3\textwidth}
% \centering
% \includegraphics[width=5cm]{figure/projection/tsne_update/tsne_force_in_domain_gan_100000.png}
% \end{minipage}
% \caption{Projection results using tsne method. The first image is the original projection results of Zhou's latest model; the second image is the projection results without outliers of Zhou's latest model; the last image is the projection result of our model. For both figure, "ori" means the true distribution of intermediate latent codes and "enc" means the reconstructing intermediate latent codes.}
% \label{fig:tsne_update}
% \end{figure}
\begin{figure}[ht]
    \centering
    \includegraphics[width=0.5\textwidth]{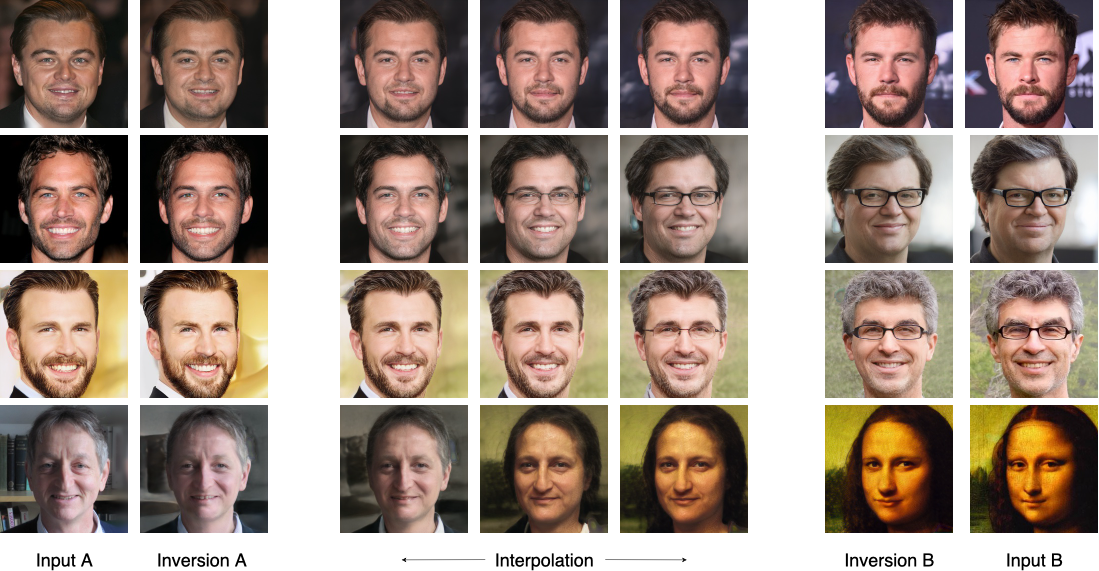}
    \caption{Image interpolation results using our \textit{force-in-domian} GAN.}
    \label{fig:interpolation}
\end{figure}
\subsection{Real Image Editing}
In this section, we apply \textit{force-in-domain} GAN inversion approach in real image editing tasks, including image interpolation, semantic diffusion and semantic manipulation.
\\{\bfseries Image Interpolation} For given two images, we can interpolate them similar to the  interpolation between two points. Image interpolation aims at semantically interpolating two images but not simply on the pixel level. Thus, it is appropriate to interpolate two images through their latent codes. It is reasonably to expect that the semantic of the interpolated image would vary continuously with the weights of two inverted codes. As shown in  Fig. \ref{fig:interpolation}, for the two input images (first and the last column), we use  \textit{force-in-domain} GAN to obtain the inversion images. Then, we interpolate the inverted latent codes and feed the interpolated latent codes into the generator. A smooth variation of the interpolated images can be clearly achieved by the \textit{force-in-domain} GAN.
% \\{\bfseries Style Mixing}
\begin{figure}[ht]
    \centering
    \includegraphics[width=0.45\textwidth]{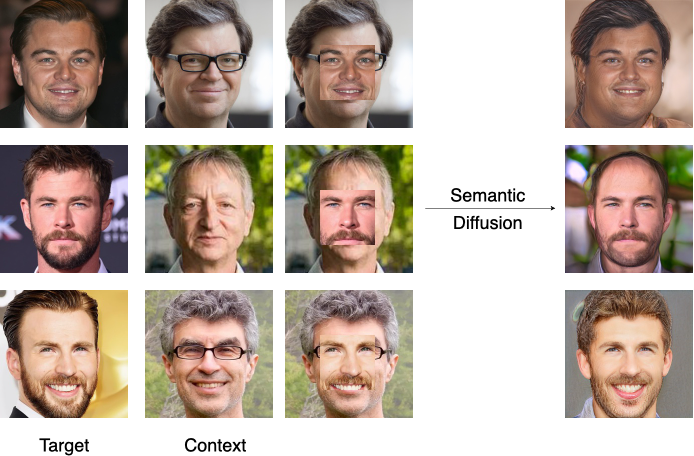}
    \caption{Semantic diffusion result using our \textit{force-in-domain} GAN.}
    \label{fig:diffusion}
\end{figure}
\\{\bfseries Semantic Diffusion.}
Semantic diffusion is a task that  diffuses a representative part of the target image into the context of another image. In our task, we diffuse the center part of a face image to another face image. As shown in   Fig. \ref{fig:diffusion}, we mask the context face (second column) by the center of the target face (first column). We obtain the inverted latent code of the mixed face through \textit{force-in-domain} GAN and reconstruct the mixed face through the generator. As shown in the last column, the reconstructed faces well preserve the identity of the target fact and  reasonably integrate into different surroundings.    
\begin{figure*}[hb]
    \centering
    \includegraphics[width=0.7\textwidth]{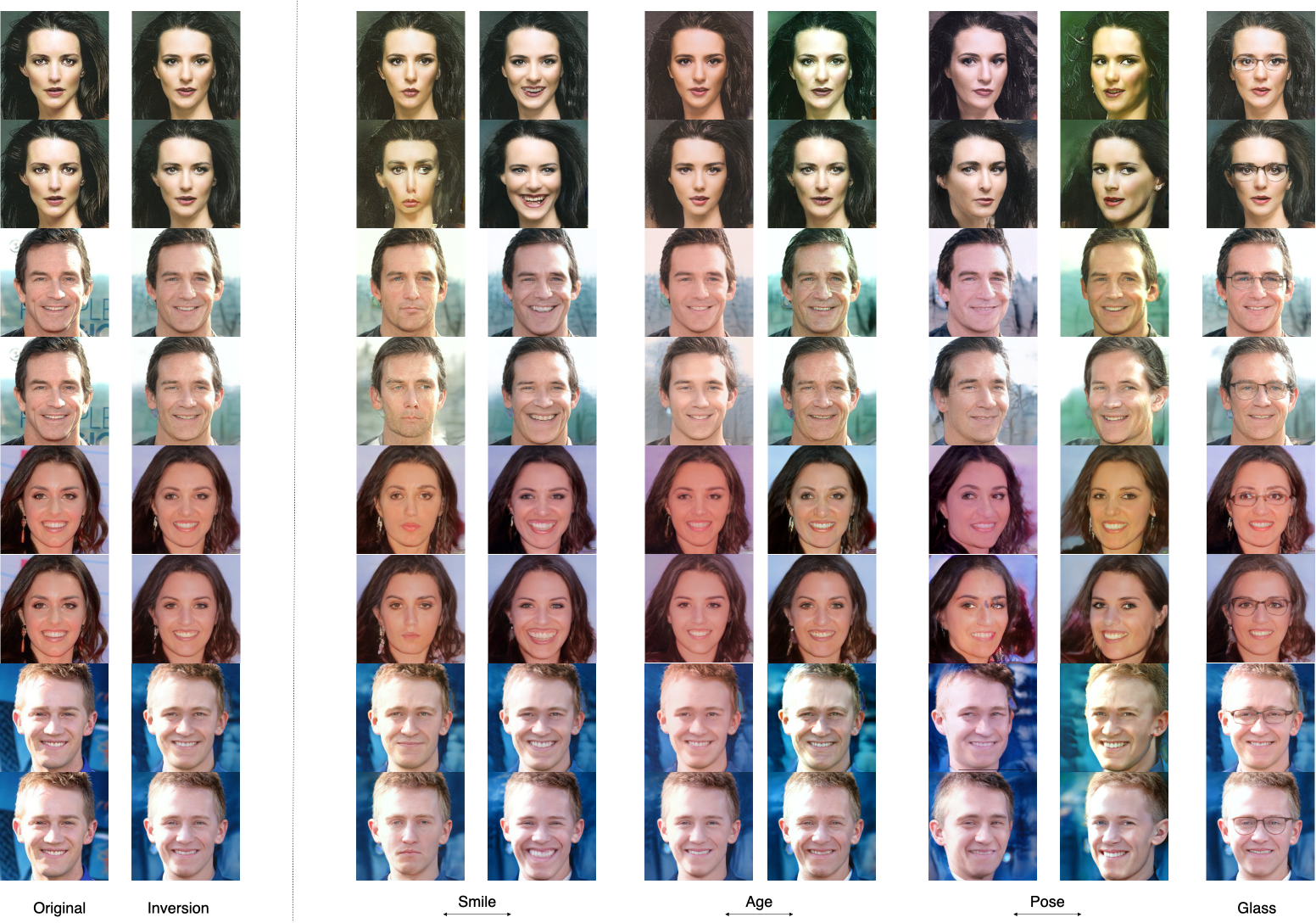}
    \caption{Semantic manipulation result using our \textit{force-in-domain} GAN.}
    \label{fig:semantic}
\end{figure*}
\\{\bfseries Semantic Manipulation.} An important and interesting task for GAN inversion is  image manipulation in semantic level. As the latent code encodes rich semantic knowledge during the training, it is able to find decision boundaries for different semantic attributes, by which we can manipulate the image through tuning the latent code towards or away from the boundary. Following the procedure in \citep{shen2020interfacegan}, we first use a pre-trained attribute prediction model  \footnote{\url{https://www.faceplusplus.com.cn}} to assign attribute scores for synthesized images. Then, for 
% is another way to examine whether the embedded latent codes align with the semantic knowledge learned by GANs. As pointed out by prior work, GANs can learn rich semantic latent representation. 
% For a given pre-trained GAN model, we can first use the pre-trained attribute prediction model to assign attributes scores for synthesized images. 
% For 
each attribute, we  learn a linear SVM, resulting in a decision boundary. Then the image editing process can be formulated as 
\begin{equation}\nonumber
\mathbf{x}^{edit} = G(\mathbf{z}^{inv} + \alpha \mathbf{n}),
\end{equation}
where $\mathbf{n}$ is the normal direction of the decision boundary for a particular semantic attribute and $\alpha$ is the step size for manipulation. It is also expected to see a continuous variation of the edited image as the step size $\alpha$. As shown in Fig. \ref{fig:semantic}, we manipulate real images on the semantics of smile, age, pose and glass based on the \textit{force-in-domain} GAN inversion. For different examples, we obtain satisfying manipulation results.
% \\{\bf Fig.\ref{fig:semantic}} show the comparison results of manipulating faces using \textit{in-domain} GAN and our \textit{force-in-domain} GAN inversion. We can see that our method shows more satisfying manipulation results than \textit{in-domain} GAN. 
% \begin{landscape}
% \begin{figure}[!ht]
%     \centering
%     \includegraphics[height=15cm]{figure/semantic/semantic.png}
%     \caption{Our \textit{force-in-doamin} GAN inversion on facial attribute manipulation, for every image, the first row to the second row means the intermediate latent codes move along the semantic direction}
%     \label{fig:my_label}
% \end{figure}
% \end{landscape}

\section{Conclusion \label{sec:conclusion}}
A key factor that affects the quality of GAN inversion is whether the inverted code lies in the original latent space. In this work, we have proposed a \textit{force-in-domain} GAN that can faithfully align the inverted codes with original latent space, thus, being a good model for real image editing.

%---------------------- 主体结束 ---------------------------------

\section{Acknowledgements}
 We thank Prof. Bolei Zhou and Prof. Lei Zhang for helpful discussions.

% \section{Acknowledgements}
% Z.X. is supported by National Key R\&D Program of China (2019YFA0709503), Shanghai Sailing Program, Natural Science Foundation of Shanghai (20ZR1429000), NSFC 62002221 and partially supported by HPC of School of Mathematical Sciences and Student Innovation Center at Shanghai Jiao Tong University.

% \bibliographystyle{aaai21}
\bibliography{DLRef}

\begin{thebibliography}{27}
\providecommand{\natexlab}[1]{#1}
\providecommand{\url}[1]{\texttt{#1}}
\providecommand{\urlprefix}{URL }
\expandafter\ifx\csname urlstyle\endcsname\relax
  \providecommand{\doi}[1]{doi:\discretionary{}{}{}#1}\else
  \providecommand{\doi}{doi:\discretionary{}{}{}\begingroup
  \urlstyle{rm}\Url}\fi

\bibitem[{Abdal, Qin, and Wonka(2019)}]{abdal2019image2stylegan}
Abdal, R.; Qin, Y.; and Wonka, P. 2019.
\newblock Image2stylegan: How to embed images into the stylegan latent space?
\newblock In \emph{Proceedings of the IEEE/CVF International Conference on
  Computer Vision}, 4432--4441.

\bibitem[{Abdal, Qin, and Wonka(2020)}]{abdal2020image2stylegan++}
Abdal, R.; Qin, Y.; and Wonka, P. 2020.
\newblock Image2stylegan++: How to edit the embedded images?
\newblock In \emph{Proceedings of the IEEE/CVF Conference on Computer Vision
  and Pattern Recognition}, 8296--8305.

\bibitem[{Arvanitidis, Hansen, and Hauberg(2017)}]{arvanitidis2017latent}
Arvanitidis, G.; Hansen, L.~K.; and Hauberg, S. 2017.
\newblock Latent space oddity: on the curvature of deep generative models.
\newblock \emph{arXiv preprint arXiv:1710.11379} .

\bibitem[{Chen et~al.(2018)Chen, Klushyn, Kurle, Jiang, Bayer, and
  Smagt}]{chen2018metrics}
Chen, N.; Klushyn, A.; Kurle, R.; Jiang, X.; Bayer, J.; and Smagt, P. 2018.
\newblock Metrics for deep generative models.
\newblock In \emph{International Conference on Artificial Intelligence and
  Statistics}, 1540--1550. PMLR.

\bibitem[{Chen et~al.(2016)Chen, Duan, Houthooft, Schulman, Sutskever, and
  Abbeel}]{chen2016infogan}
Chen, X.; Duan, Y.; Houthooft, R.; Schulman, J.; Sutskever, I.; and Abbeel, P.
  2016.
\newblock Infogan: Interpretable representation learning by information
  maximizing generative adversarial nets.
\newblock \emph{arXiv preprint arXiv:1606.03657} .

\bibitem[{Creswell and Bharath(2018)}]{creswell2018inverting}
Creswell, A.; and Bharath, A.~A. 2018.
\newblock Inverting the generator of a generative adversarial network.
\newblock \emph{IEEE transactions on neural networks and learning systems}
  30(7): 1967--1974.

\bibitem[{Donahue et~al.(2017)Donahue, Lipton, Balsubramani, and
  McAuley}]{donahue2017semantically}
Donahue, C.; Lipton, Z.~C.; Balsubramani, A.; and McAuley, J. 2017.
\newblock Semantically decomposing the latent spaces of generative adversarial
  networks.
\newblock \emph{arXiv preprint arXiv:1705.07904} .

\bibitem[{Donahue, Kr{\"a}henb{\"u}hl, and
  Darrell(2016)}]{donahue2016adversarial}
Donahue, J.; Kr{\"a}henb{\"u}hl, P.; and Darrell, T. 2016.
\newblock Adversarial feature learning.
\newblock \emph{arXiv preprint arXiv:1605.09782} .

\bibitem[{Dumoulin et~al.(2016)Dumoulin, Belghazi, Poole, Mastropietro, Lamb,
  Arjovsky, and Courville}]{dumoulin2016adversarially}
Dumoulin, V.; Belghazi, I.; Poole, B.; Mastropietro, O.; Lamb, A.; Arjovsky,
  M.; and Courville, A. 2016.
\newblock Adversarially learned inference.
\newblock \emph{arXiv preprint arXiv:1606.00704} .

\bibitem[{Goetschalckx et~al.(2019)Goetschalckx, Andonian, Oliva, and
  Isola}]{goetschalckx2019ganalyze}
Goetschalckx, L.; Andonian, A.; Oliva, A.; and Isola, P. 2019.
\newblock Ganalyze: Toward visual definitions of cognitive image properties.
\newblock In \emph{Proceedings of the IEEE/CVF International Conference on
  Computer Vision}, 5744--5753.

\bibitem[{Goodfellow et~al.(2014)Goodfellow, Pouget-Abadie, Mirza, Xu,
  Warde-Farley, Ozair, Courville, and Bengio}]{goodfellow2014generative}
Goodfellow, I.~J.; Pouget-Abadie, J.; Mirza, M.; Xu, B.; Warde-Farley, D.;
  Ozair, S.; Courville, A.; and Bengio, Y. 2014.
\newblock Generative adversarial networks.
\newblock \emph{arXiv preprint arXiv:1406.2661} .

\bibitem[{Heusel et~al.(2017)Heusel, Ramsauer, Unterthiner, Nessler, Klambauer,
  and Hochreiter}]{heusel2017gans}
Heusel, M.; Ramsauer, H.; Unterthiner, T.; Nessler, B.; Klambauer, G.; and
  Hochreiter, S. 2017.
\newblock GANs Trained by a Two Time-Scale Update Rule Converge to a Nash
  Equilibrium. .

\bibitem[{Jahanian, Chai, and Isola(2019)}]{jahanian2019steerability}
Jahanian, A.; Chai, L.; and Isola, P. 2019.
\newblock On the" steerability" of generative adversarial networks.
\newblock \emph{arXiv preprint arXiv:1907.07171} .

\bibitem[{Johnson, Alahi, and Fei-Fei(2016)}]{johnson2016perceptual}
Johnson, J.; Alahi, A.; and Fei-Fei, L. 2016.
\newblock Perceptual losses for real-time style transfer and super-resolution.
\newblock In \emph{European conference on computer vision}, 694--711. Springer.

\bibitem[{Karras, Laine, and Aila(2019)}]{karras2019style}
Karras, T.; Laine, S.; and Aila, T. 2019.
\newblock A style-based generator architecture for generative adversarial
  networks.
\newblock In \emph{Proceedings of the IEEE/CVF Conference on Computer Vision
  and Pattern Recognition}, 4401--4410.

\bibitem[{Kuhnel et~al.(2018)Kuhnel, Fletcher, Joshi, and
  Sommer}]{kuhnel2018latent}
Kuhnel, L.; Fletcher, T.; Joshi, S.; and Sommer, S. 2018.
\newblock Latent space non-linear statistics.
\newblock \emph{arXiv preprint arXiv:1805.07632} .

\bibitem[{Lipton and Tripathi(2017)}]{lipton2017precise}
Lipton, Z.~C.; and Tripathi, S. 2017.
\newblock Precise recovery of latent vectors from generative adversarial
  networks.
\newblock \emph{arXiv preprint arXiv:1702.04782} .

\bibitem[{Ma, Ayaz, and Karaman(2019)}]{ma2019invertibility}
Ma, F.; Ayaz, U.; and Karaman, S. 2019.
\newblock Invertibility of convolutional generative networks from partial
  measurements .

\bibitem[{Odena, Olah, and Shlens(2017)}]{odena2017conditional}
Odena, A.; Olah, C.; and Shlens, J. 2017.
\newblock Conditional image synthesis with auxiliary classifier gans.
\newblock In \emph{International conference on machine learning}, 2642--2651.
  PMLR.

\bibitem[{Perarnau et~al.(2016)Perarnau, Van De~Weijer, Raducanu, and
  {\'A}lvarez}]{perarnau2016invertible}
Perarnau, G.; Van De~Weijer, J.; Raducanu, B.; and {\'A}lvarez, J.~M. 2016.
\newblock Invertible conditional gans for image editing.
\newblock \emph{arXiv preprint arXiv:1611.06355} .

\bibitem[{Shen et~al.(2018)Shen, Luo, Yan, Wang, and Tang}]{shen2018faceid}
Shen, Y.; Luo, P.; Yan, J.; Wang, X.; and Tang, X. 2018.
\newblock Faceid-gan: Learning a symmetry three-player gan for
  identity-preserving face synthesis.
\newblock In \emph{Proceedings of the IEEE conference on computer vision and
  pattern recognition}, 821--830.

\bibitem[{Shen et~al.(2020)Shen, Yang, Tang, and Zhou}]{shen2020interfacegan}
Shen, Y.; Yang, C.; Tang, X.; and Zhou, B. 2020.
\newblock Interfacegan: Interpreting the disentangled face representation
  learned by gans.
\newblock \emph{IEEE Transactions on Pattern Analysis and Machine Intelligence}
  .

\bibitem[{Simonyan and Zisserman(2014)}]{simonyan2014very}
Simonyan, K.; and Zisserman, A. 2014.
\newblock Very deep convolutional networks for large-scale image recognition.
\newblock \emph{arXiv preprint arXiv:1409.1556} .

\bibitem[{Tran, Yin, and Liu(2017)}]{tran2017disentangled}
Tran, L.; Yin, X.; and Liu, X. 2017.
\newblock Disentangled representation learning gan for pose-invariant face
  recognition.
\newblock In \emph{Proceedings of the IEEE conference on computer vision and
  pattern recognition}, 1415--1424.

\bibitem[{Van~der Maaten and Hinton(2008)}]{van2008visualizing}
Van~der Maaten, L.; and Hinton, G. 2008.
\newblock Visualizing data using t-SNE.
\newblock \emph{Journal of machine learning research} 9(11).

\bibitem[{Zhu et~al.(2020)Zhu, Shen, Zhao, and Zhou}]{zhu2020domain}
Zhu, J.; Shen, Y.; Zhao, D.; and Zhou, B. 2020.
\newblock In-domain gan inversion for real image editing.
\newblock In \emph{European Conference on Computer Vision}, 592--608. Springer.

\bibitem[{Zhu et~al.(2019)Zhu, Zhao, Zhou, and Zhang}]{zhu2019lia}
Zhu, J.; Zhao, D.; Zhou, B.; and Zhang, B. 2019.
\newblock Lia: Latently invertible autoencoder with adversarial learning .

\end{thebibliography}

\end{document}